\documentclass[10pt,twocolumn,letterpaper]{article}

\usepackage[pagenumbers]{cvpr}      

\usepackage[most]{tcolorbox}
\usepackage{placeins}
\usepackage{graphicx}
\definecolor{cvprblue}{rgb}{0.21,0.49,0.74}
\usepackage[pagebackref,breaklinks,colorlinks,allcolors=cvprblue]{hyperref}

\title{Locatability-Guided Adaptive Reasoning for Image Geo-Localization with Vision-Language Models}

\author{
    Bo Yu\textsuperscript{1} \quad
    Fengze Yang\textsuperscript{1} \quad
    Yiming Liu\textsuperscript{2} \quad
    Chao Wang\textsuperscript{3} \quad
    Xuewen Luo\textsuperscript{1} \\
    Taozhe Li\textsuperscript{2} \quad
    Ruimin Ke\textsuperscript{4} \quad
    Xiaofan Zhou\textsuperscript{5} \quad
    Chenxi Liu\textsuperscript{1*} \\[2mm] 
    \textsuperscript{1}The University of Utah \qquad
    \textsuperscript{2}The University of Oklahoma \\
    \textsuperscript{3}Worcester Polytechnic Institute \qquad
    \textsuperscript{4}Rensselaer Polytechnic Institute \\
    \textsuperscript{5}University of Illinois Chicago \\[1.5mm]
    {\tt\small bo.yu@utah.edu, chenxi.liu@utah.edu}
}

\begin{document}
\maketitle
\begin{abstract}

The emergence of Vision-Language Models (VLMs) has introduced new paradigms for global image geo-localization through retrieval-augmented generation (RAG) and reasoning-driven inference. However, RAG methods are constrained by retrieval database quality, while reasoning-driven approaches fail to internalize image locatability, relying on inefficient, fixed-depth reasoning paths that increase hallucinations and degrade accuracy. To overcome these limitations, we introduce an Optimized Locatability Score that quantifies an image's suitability for deep reasoning in geo-localization. Using this metric, we curate Geo-ADAPT-51K, a locatability-stratified reasoning dataset enriched with augmented reasoning trajectories for complex visual scenes. Building on this foundation, we propose a two-stage Group Relative Policy Optimization (GRPO) curriculum with customized reward functions that regulate adaptive reasoning depth, visual grounding, and hierarchical geographical accuracy. Our framework, Geo-ADAPT, learns an adaptive reasoning policy, achieves state-of-the-art performance across multiple geo-localization benchmarks, and substantially reduces hallucinations by reasoning both adaptively and efficiently.

\end{abstract}    
\section{Introduction}
\label{sec:intro}

Global image geo-localization \cite{Vo17} refers to the task of estimating the geographic coordinates of a scene from a single image captured at any location on Earth. It is a core problem in computer vision with broad applications in autonomous navigation, social media analysis, disaster management, and environmental monitoring \cite{durgam2024cross, li2025cross, savarro2024leveraging}. In contrast to region-specific methods \cite{durgam2024cross}, global-scale image localization must address the extensive geographic range and visual heterogeneity of the entire planet, a task made challenging by significant visual ambiguity and large intra-class variation. Traditional approaches are typically categorized into classification-based methods \cite{weyand2016planet, seo2018cplanet, pramanick2022world, clark2023we, haas2024pigeon}, which suffer from a fundamental precision-versus-scalability trade-off, and retrieval-based methods \cite{zhu2022transgeo, lin2022joint, zhang2023cross, workman2015wide, liu2019lending, zhu2021vigor}, which are constrained by the cost, density, and coverage of their required geo-referenced databases.

\begin{figure}
    \centering
    \includegraphics[width=1\linewidth]{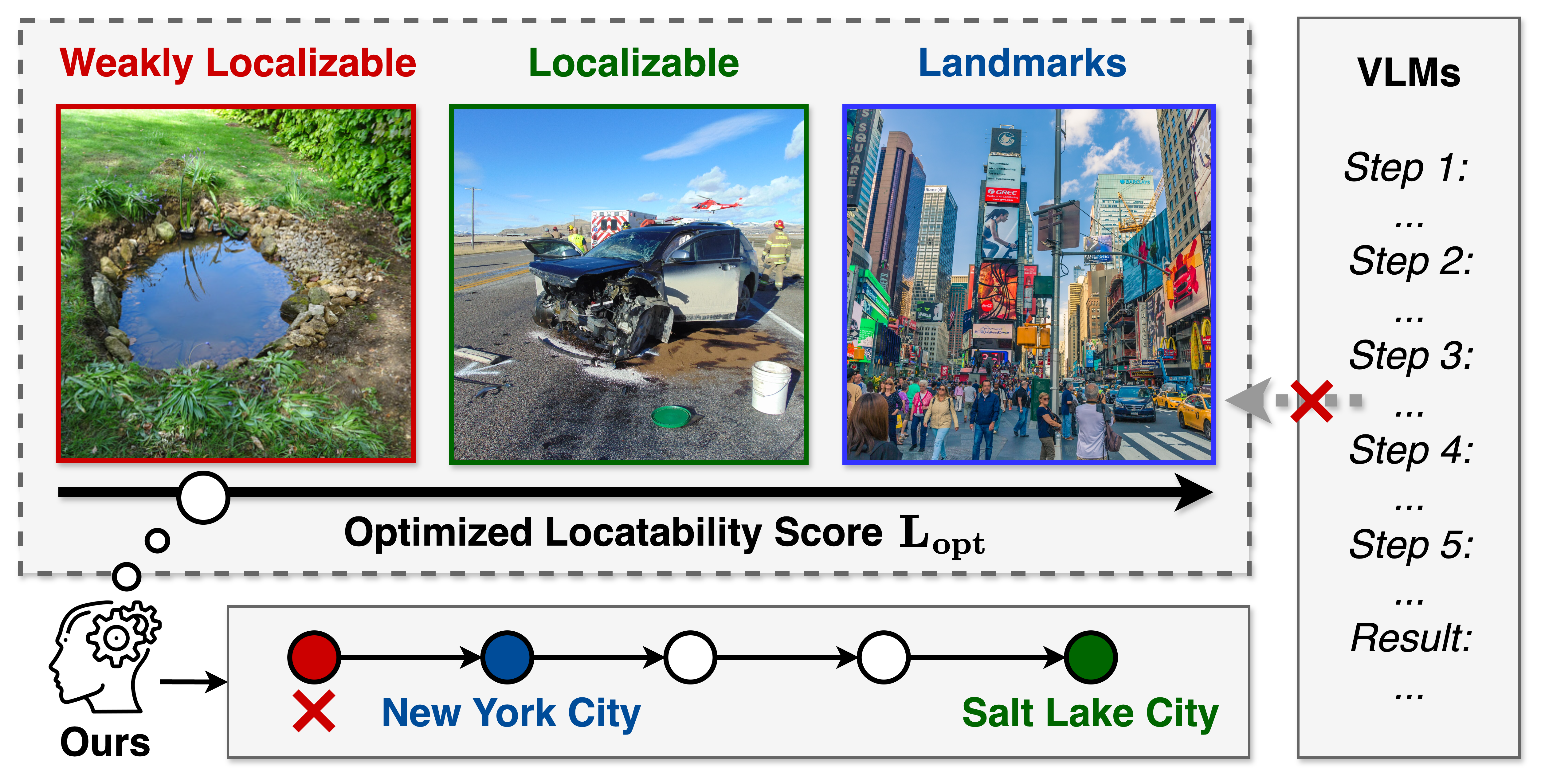}
    \caption{Illustration of locatability-guided adaptive reasoning. Current reasoning VLMs for image geo-localization are trained on fixed chain of thoughts, generating predetermined reasoning trajectories step by step without considering the actual image locatability during inference. In our work, we propose Geo-ADAPT, an adaptive reasoning framework that dynamically adjusts reasoning depth based on the image's Optimized Locatability Score $L_{opt}$.}
    \label{fig:image_1}
\end{figure}

The emergence of VLMs \cite{li2024llava, wang2024qwen2, bordes2024introductionvisionlanguagemodeling} has introduced a new paradigm for image geo-localization, branching into two primary avenues: RAG \cite{Jia24G3, zhou2024img2loc, jia2025georanker, lewis2020retrieval} and reasoning-driven inference \cite{Li24Georeasoner, Wang25GRESuite, Li25Recognition}. RAG, while effective, remains reliant on the quality of its retrieval databases and does not fully leverage the advanced reasoning capabilities of VLMs. In contrast, reasoning-driven methods aim to emulate the multi-step cognitive process of image geo-localization \cite{Li25Recognition, Wang25GRESuite}. However, they face a critical flaw: they fail to internalize a robust concept of image locatability in the reasoning process \cite{Li24Georeasoner}, which quantifies how visually identifiable a location is, resorting to inefficient, fixed-depth reasoning paths. This ``one-size-fits-all'' policy applies the same inferential depth to simple and complex images alike, increasing hallucinations and degrading accuracy. Figure \ref{fig:image_1} illustrates the limitations of current VLMs reasoning for image geo-localization and our proposed solution.

To address these challenges, we propose \textbf{Geo-ADAPT}, a novel framework that learns an adaptive reasoning policy for image geo-localization. Our approach is three-fold: First, we introduce an Optimized Locatability Score $L_{opt}$, a metric that differentiates an image that is truly non-localizable from one that simply requires multi-step, implicit inference (deep reasoning) to solve. Second, using this metric, we curate \textbf{Geo-ADAPT-51K}, a new 51K-image dataset from IMAGEO-Bench \cite{Li25PixelsToPlaces} and X (formerly Twitter). This dataset is uniquely stratified by the Optimized Locatability Score $L_{opt}$ and enriched with augmented, implicit reasoning trajectories for complex scenes requiring deep reasoning. Third, we propose a two-stage GRPO \cite{shao2024deepseekmath}-based curriculum built on tailored reward functions governing adaptive reasoning depth, visual grounding, and hierarchical geographical accuracy. Specifically, this curriculum trains the policy in two phases: Stage 1 (Reasoning Formation) focuses on learning how and when to reason deeply and in a visually-grounded manner. Stage 2 (Localization Refinement) then optimizes for precise, hierarchical geo-prediction accuracy, while a KL penalty preserves the foundational reasoning skills. The resulting model \textbf{Geo-ADAPT-8B} achieves state-of-the-art performance, significantly reducing hallucinations by reasoning adaptively, accurately, and efficiently. To summarize, our contributions are as follows:
\vspace{-1mm}
\begin{itemize}[topsep=0pt, itemsep=0pt, parsep=0pt]
    \item We propose an Optimized Locatability Score $L_{opt}$ that quantifies real reasoning feasibility for image geo-localization, forming the basis for stratified data curation, customized reward construction, and GRPO training.

    \item We curate Geo-ADAPT-51K, a 51K-image dataset stratified by the Optimized Locatability Score $L_{opt}$ and enriched with implicit reasoning trajectories for deep reasoning, the first dataset designed for training adaptive reasoning policies in image geo-localization.
    
    \item We develop a two-stage GRPO curriculum that learns adaptive policies by first optimizing reasoning formation, then refining hierarchical localization, achieving state-of-the-art performance across multiple benchmarks.
\end{itemize}


\section{Related Work}
\label{sec:Related}

\subsection{Traditional Methods}
Image geo-localization is a challenging task in computer vision \cite{Vo17}, GeoAI \cite{mai2024opportunities, janowicz2020geoai, Li24Georeasoner}, and spatial data mining \cite{wang2020deep, liang2025foundation}. Traditional approaches have been dominated by two paradigms. The first, classification-based methods \cite{weyand2016planet, seo2018cplanet, pramanick2022world, clark2023we, haas2024pigeon}, partitions the Earth's surface into a discrete grid, creating an inflexible trade-off between localization precision and output space scalability. The second, retrieval-based methods \cite{zhu2022transgeo, lin2022joint, zhang2023cross, workman2015wide, liu2019lending, zhu2021vigor, vivanco2023geoclip}, matches a query against a large geo-referenced database but is fundamentally constrained by the database's cost, density, and coverage. The advent of VLMs, however, has introduced a paradigm shift, offering new potential to overcome the limitations of these traditional approaches.

\subsection{Vision Language Models}
VLMs are variants of Large Language Models (LLMs). In addition to processing and understanding text, they also have the advanced ability to process visual information. Their evolution progressed from contrastive learning \cite{li2022blip, Radford21CLIP} and early multimodal analysis \cite{alayrac2022flamingo} to sophisticated instruction-tuned systems \cite{dai2023instructblip, liu2023visual}. Recent models like GPT-4o \cite{gpt4o} and other advanced VLMs \cite{li2024llava, liu2024deepseek, wang2024qwen2, zhang2024internlm} now exhibit powerful multimodal reasoning \cite{bordes2024introductionvisionlanguagemodeling}, which has spurred two distinct approaches in image geo-localization: RAG and reasoning-driven inference. 

RAG-based methods \cite{Jia24G3, zhou2024img2loc, jia2025georanker} provide VLMs with external image candidates with extra information such as brief text and coordinates. However, their performance remains fundamentally bottlenecked by the quality and coverage of the reference datasets, and they under-utilize the VLM's advanced reasoning capabilities. Conversely, reasoning-driven inference \cite{Li24Georeasoner, Wang25GRESuite, Li25Recognition} aims to maximize the VLM's internal prior knowledge, often refined via Supervised Fine-Tuning (SFT) or Reinforcement Learning (RL), to perform localization without an external database. These methods, however, suffer from a critical flaw: they employ inefficient, ``one-size-fits-all" reasoning policies. By failing to internalize image locatability, they apply a fixed reasoning depth to all images, which increases hallucinations and degrades accuracy. Thus, there is an urgent need for adaptive reasoning methods to address these limitations. 

\subsection{Reinforcement Learning} RL is one of the candidates that has the greatest potential to solve this problem. Recently, RL has proven highly effective for enhancing complex reasoning in the text domain, with notable successes in tasks like mathematics \cite{cai2024internlm2, shao2024deepseekmath, wang2024qwen2, ying2024internlm} and coding \cite{hui2024qwen2, jiao2024preference, zhang2025codedpo, zhang2024o1codero1replicationcoding, schulman2017proximal}. Breakthroughs such as DeepSeek-R1 \cite{deepseekai2025deepseekr1incentivizingreasoningcapability}, which utilize GRPO \cite{shao2024deepseekmath}, have demonstrated that RL can instill robust reasoning abilities, even replacing conventional SFT stages. While RL has become a standard tool for enhancing the reasoning and instruction-following abilities of LLMs \cite{abdulhai2023lmrl, carta2023grounding, ouyang2022training, ramamurthy2022reinforcement, snell2022offline, sun2024aligning, yao2022react, zang2025contextual, zhou2024archer, ziegler2019fine}, its application to the unique challenges of multimodal reasoning in VLMs is a more nascent research area. In the multimodal domain, however, RL research has been less focused on core reasoning. Instead, it is primarily employed to mitigate hallucinations or align VLMs with human preferences \cite{liu2024miadpomultiimageaugmenteddirect, sun2024aligning, DBLP:conf/cvpr/YuYZHHCHL0024, yu2025rlaifvopensourceaifeedback, zhao2024hallucinationsenhancinglvlmshallucinationaware, zhou2024aligningmodalitiesvisionlarge}. A significant gap persists in applying RL to enhance the visual and spatial reasoning of VLMs for complex tasks. 

Our work addresses this gap by introducing a novel, two-stage GRPO \cite{shao2024deepseekmath} curriculum for the image geo-localization task. We apply a suite of verifiable, customized task-specific rewards to train an adaptive policy, improving the VLM's ability to process diverse inputs and reason about visual-spatial evidence. Additionally, our work leverages VLMs within a locatability-guided, adaptive, retrieval-free framework that internalizes an Optimized Locatability Score, enabling the model to dynamically allocate its reasoning depth based on the scene's difficulty.

\section{Methodology}
\label{sec:methodology}

\begin{figure*}[t]
    \centering
    \includegraphics[width=1\linewidth]{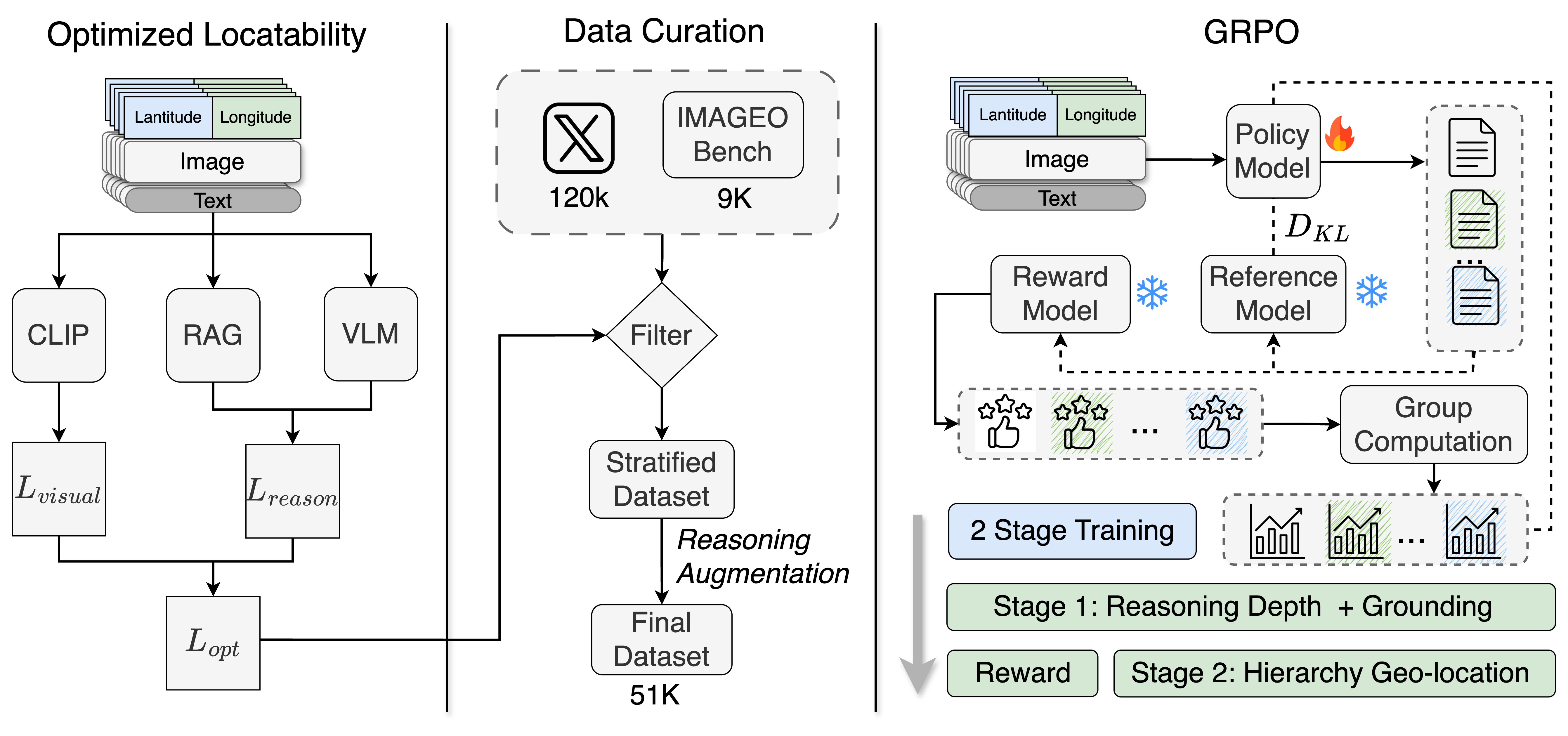}
    \caption{Overview of Geo-ADAPT framework. Our approach comprises three components: (1) an Optimized Locatability Score $L_{opt}$ quantifying reasoning feasibility, (2) locatability-stratified dataset curation Geo-ADAPT-51K with enriched implicit reasoning trajectories, and (3) two-stage GRPO training with adaptive depth, grounding, and hierarchical accuracy rewards for dynamic reasoning allocation.}
    \label{fig:image_2}
\end{figure*}

Given an image $I$ captured at an unknown location, the goal of image geo-localization is to predict its geographical location $g = (c, t, coord)$, where $c$ denotes the country, $t$ the city, and $coord$ the GPS coordinates. Unlike traditional classification or retrieval-based approaches, reasoning-driven geo-localization aims to generate interpretable reasoning trajectories $r = \{s_1, s_2, \ldots, s_n\}$ that articulate the visual cues and geographical knowledge leading to the prediction. However, current reasoning methods apply fixed-depth inference uniformly across all images with pre-designed chain of thoughts. To address this challenge, we propose a novel framework, Geo-ADAPT, illustrated in Figure \ref{fig:image_2}. The following subsections detail each component: Optimized Locatability Score (Section \ref{sec:localibity_score}), data curation (Section \ref{sec:data_curation}), reward construction (Section \ref{sec:reward_construction}), and GRPO-based RL (Section \ref{sec:grpo_rl}).

\subsection{Optimized Locatability Score}
\label{sec:localibity_score}

The original locatability score for a given image \cite{Li24Georeasoner}, ranging from 0 to 1, quantifies how visually identifiable a location is, with higher scores indicating more distinctive scenes. The score is computed by taking the area ratios of MaskFormer segmentation classes \cite{Cheng21MaskFormer}, deriving importance weights from CLIP- and Sentence-BERT–based text–label similarities \cite{Li24Georeasoner, Radford21CLIP, Reimers19SentenceBERT}, normalizing them, and then summing the class ratios weighted by these learned importance values.
However, this metric exhibits two limitations: (1) it serves primarily as a post-hoc analysis or data curation tool, rather than as a dynamic signal integrated into a model's reasoning inference; and (2) this metric is designed for SFT, is static and superficial, failing to assess an image's amenability to deep reasoning. That means the metric cannot differentiate an image that is truly non-localizable from one that simply requires multi-step, implicit inference (deep reasoning) to solve. To address these shortcomings, we propose the Optimized Locatability Score that explicitly quantifies an image's amenability to deep, multi-step reasoning, serving as a dynamic signal integrated into the model's policy rather than as a post-hoc analysis tool for data curation.

\textbf{Reasoning Locatability Score.}
While $L_{visual}$ captures explicit, static cues, it fails to address the limitations we identified previously. 
To quantify this potential for image geo-localization by leveraging deep, multi-step inference, inspired by DeepSeek-R1 \cite{Guo25DeepSeek}, we introduce the Reasoning Locatability Score $L_{reason}$ to optimize the Original Locatability Score $L_{visual}$ for image geo-localization.

\begin{figure}[t]
    \centering
    \includegraphics[width=0.95\linewidth]{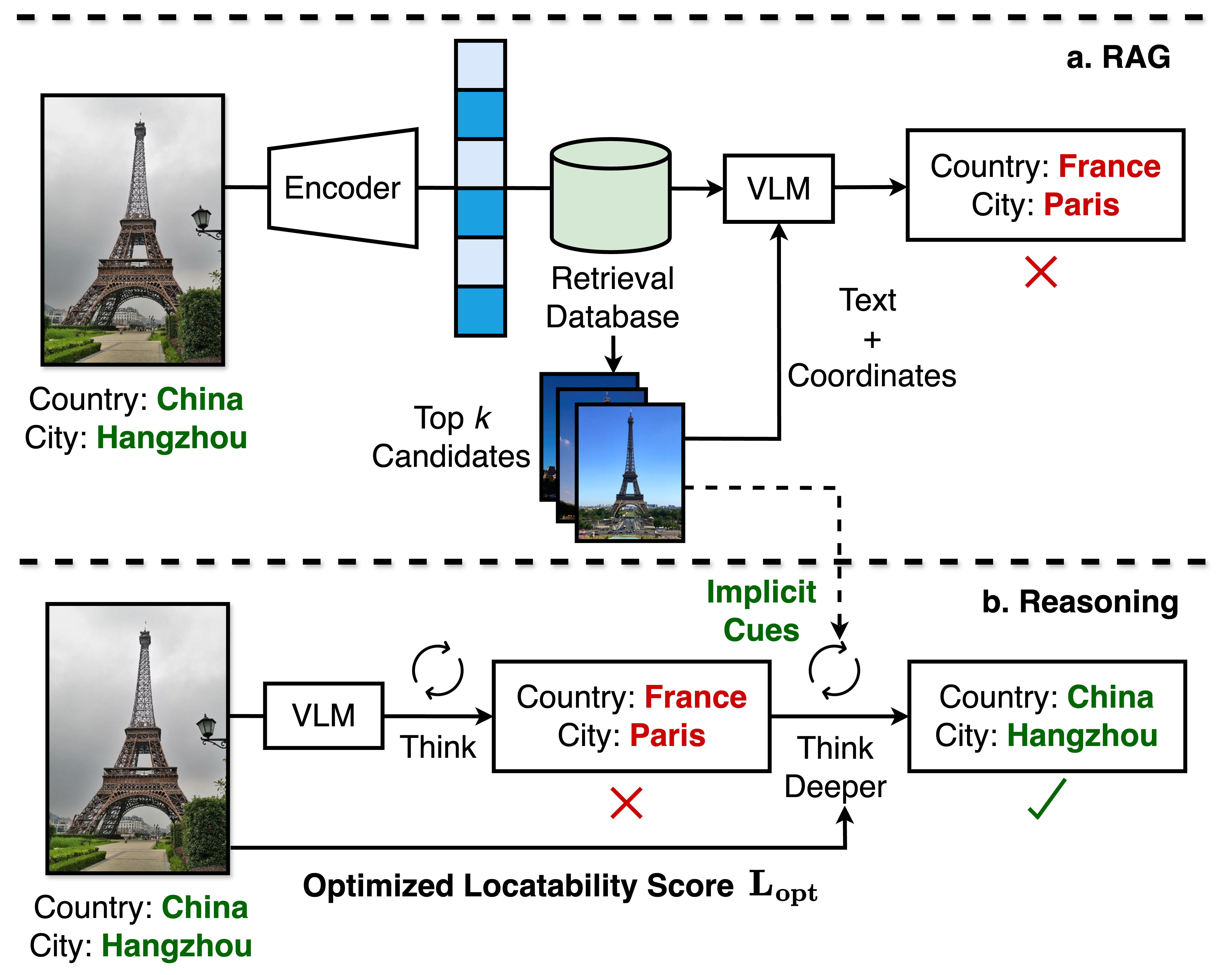}
    \caption{Motivation for Optimized Locatability Score $L_{opt}$ and locatability-guided adaptive reasoning strategies. (a) RAG retrieves visually similar candidates but fails due to being unable to convert implicit visual cues to rich semantic priors. (b) Standard reasoning initially fails similarly, but integrating implicit cues from retrieved image candidates into deep reasoning enables correct localization. Images where $d_{Reason} > d_{RAG} + \tau_{margin}$ receive lower $L_{opt}$, triggering adaptive deep reasoning.}
    \label{fig:image_3}
\end{figure}

The design of $L_{reason}$ stems from a key observation based on two mainstream VLM-based image geo-localization paradigms: RAG and reasoning inference. Reasoning-driven VLMs excel at localizing images with explicit visual cues (e.g., landmarks, signage), while RAG methods leverage retrieved candidates to better handle images with subtle, implicit patterns. However, current RAG approaches fail to convert these implicit visual cues from retrieved candidates into rich semantic priors that can guide VLM reasoning. Through experiments, we find that extracting implicit cues from RAG-retrieved top-$k$ image candidates and integrating them into reasoning chains substantially enhances geo-localization performance for a large number of images. This motivates our $L_{reason}$ formulation: we quantify reasoning locatability via the performance gap between RAG and reasoning approaches. 
Images where RAG significantly outperforms reasoning indicate learnable implicit patterns that benefit from enriched reasoning trajectories, while superior or comparable VLM reasoning performance suggests sufficient explicit cues for standard reasoning. 
This idea enables systematic identification of images amenable to deep reasoning enhancement and accurately reflects reasoning-based locatability. The design motivation for the Reasoning Locatability Score $L_{reason}$ and the locatability-guided adaptive reasoning mechanism introduced in Sections \ref{sec:reward_construction} and \ref{sec:grpo_rl} is illustrated in Figure \ref{fig:image_3}.

To compute this, we first establish baseline performance with ground-truth coordinates $P_{GT}$. We generate predictions ($P_{RAG}$) from a SOTA RAG method, GeoRanker \cite{jia2025georanker}, and predictions ($P_{Reason}$) from a SOTA reasoning VLM, GRE Suite \cite{Wang25GRESuite}. We then compute their respective Haversine distances to the ground truth: $d_{RAG} = d(P_{RAG}, P_{GT})$ and $d_{Reason} = d(P_{Reason}, P_{GT})$. Thus, the Reasoning Locatability Score $L_{reason}$ is as below:
\begin{equation}
    \begin{split}
        L_{reason} &= 
        \underbrace{\exp(-\gamma_1 \cdot d_{Reason})}_{L_{base}} \\
        &\quad \cdot 
        \underbrace{\exp\left(-\gamma_2 \cdot \max(0, d_{Reason} - d_{RAG})\right)}_{L_{gap}},
    \end{split}
    \label{eq:loc_reasoning}
\end{equation}
where $\gamma_1, \gamma_2 > 0$ are scaling hyperparameters. The base term $L_{base}$ represents absolute reasoning accuracy, while the gap term $L_{gap}$ imposes an asymmetric penalty when reasoning underperforms RAG ($d_{Reason} > d_{RAG}$), suggesting that the image is better suited to RAG, is difficult to geo-localize by standard reasoning trajectories, and may require extended deep reasoning. Critically, when reasoning outperforms or matches RAG, no penalty is applied, meaning standard reasoning suffices for the given image.

\textbf{Optimized Locatability Score.}
We formulate $L_{opt}$ as a multiplicative formulation where $L_{reason}$ modulates the original visual locatability score:
\begin{equation}
L_{opt} = L_{visual} \cdot \left[(1 - \alpha) + \alpha \cdot L_{reason}\right],
\label{eq:loc_opt}
\end{equation}
where $\alpha \in [0, 1]$ controls the modulation strength. This formulation treats $L_{visual}$ as an optimistic upper bound based on explicit visual features, while $L_{reason}$ provides an empirical correction based on actual reasoning model performance. When reasoning models effectively leverage visual cues ($L_{reason} \approx 1$), minimal adjustment occurs; when they struggle ($L_{reason} \ll 1$), the score is proportionally reduced. The bounded range $L_{opt} \in [0, L_{visual}]$ ensures consistency, yielding a principled measure of reasoning-based locatability that guides model selection, data curation, and adaptive reasoning inference strategies.

\subsection{Data Curation with Optimized Locatability}
\label{sec:data_curation}

A precise computation of the Optimized Locatability Score $L_{opt}$ is critical for our adaptive reasoning framework, as this score is derived from the performance of SOTA RAG-based and reasoning inference methods as described in Section \ref{sec:localibity_score}. Instead of using the MP16-Pro dataset \cite{Jia24G3} as in previous research, which has been used to train the SOTA models, we require entirely distinct, unseen data resources to avoid biased performance metrics and the resulting inaccurate $L_{opt}$ scores from such data contamination. We therefore construct our dataset from two distinct sources: the IMAGEO-Bench \cite{Li25PixelsToPlaces} (approx. 9K images) and a novel 120K image corpus we collected from X with geo-tags (formerly Twitter) globally. This clean and unseen dataset ensures a valid and accurate calculation of the Optimized Locatability Score $L_{opt}$ for the training stage in Section \ref{sec:grpo_rl}.

\textbf{Initial Data Stratification.}
Our curation process commences by applying the Optimized Locatability Score $L_{opt}$ to the new dataset. We first generate the SOTA RAG and reasoning VLM predictions for each image, which are prerequisites for calculating $L_{opt}$ according to Equations \ref{eq:loc_reasoning} and \ref{eq:loc_opt}. This $L_{opt}$ score and its components enable us to stratify the dataset. We identify a RAG-superior subset where $d_{Reason} > d_{RAG} + \tau_{margin}$ ($\tau_{margin}$ is a distance threshold to ensure meaningful performance gaps). For this subset, we augment the data by retrieving the top-3 image candidates using the RAG method from \cite{jia2025georanker}, yielding (original, similar) image pairs. We thus divide our initial dataset into a standard image subset for standard reasoning trajectories and a RAG-superior subset for deep reasoning that leverages implicit cues from the top-3 image candidates of RAG. 
For all images, we use the method in \cite{wei2022chain} to generate reasoning trajectories and the method in \cite{Li25Recognition} for cross-verification to guarantee reasoning data quality. Through this process, we curate a dataset with 51K images.

\textbf{Reasoning Augmentation and Validation.}
For the RAG-superior image subset, in addition to standard reasoning trajectories, we also extract implicit reasoning cues from the top-3 RAG image candidates using visual grounding. Each reasoning step $s_i$ is checked by Grounding-DINO \cite{liu2024grounding} to verify visual presence. Steps with confidence below 0.3 are marked as implicit (e.g., ``architectural style suggests...''), while explicit ones (e.g., ``the sign reads...'') are removed. Implicit steps must apply to at least 2 of 3 similar images for validation. We use Gemini 2.5 Flash \cite{Comanici25Gemini2.5} to verify that this removal step is conducted correctly. The resulting implicit cues are then merged into standard reasoning chains, producing approximately 16K images with augmented reasoning trajectories.

\subsection{Reward Construction}
\label{sec:reward_construction}

Based on the dataset detailed in Section \ref{sec:data_curation}, we introduce three specialized reward functions to independently evaluate different aspects of the model's geographical reasoning. Using supervised annotations, these combined rewards provide the feedback required to guide the reinforcement learning updates described in Section \ref{sec:grpo_rl}.

Formally, let $\mathcal{D} = \{(I_i, y_i, g_i, r_i)\}_{i=1}^N$ denote the curated dataset of $N$ samples, where $I_i$ is an image, $y_i \in [0, 1]$ is the Optimized Locatability Score $L_{opt}(I_i)$, $g_i$ is the ground-truth geolocation, and $r_i$ is the reasoning trajectory.

\subsubsection{Adaptive Reasoning Depth Reward}
\label{sec:adaptive_depth_reward}

To enable adaptive reasoning patterns, we formulate a binary classification reward that guides the model to identify when deep reasoning with implicit cues is necessary. Specifically, we leverage the stratification from Section \ref{sec:data_curation}: an image belongs to the RAG-superior subset if $d_{Reason} > d_{RAG} + \tau_{margin}$, indicating that standard reasoning is insufficient and that deep reasoning is required. We define a binary ground-truth label $\ell_i$ for each image:
\begin{equation}
\ell_i = \mathbb{I}[d_{Reason}(I_i) > d_{RAG}(I_i) + \tau_{margin}],
\label{eq:rag_superior_label}
\end{equation}
where $\ell_i = 1$ indicates that the image requires elongated reasoning with implicit cue reflection, and $\ell_i = 0$ indicates that standard reasoning suffices.
The model learns to predict a reasoning depth indicator $\hat{\ell}_i \in \{0, 1\}$ based on visual features and intermediate reasoning states. The adaptive reasoning depth reward is defined as:
\begin{equation}
R_{depth}(I_i, \hat{\ell}_i) = \mathbb{I}[\hat{\ell}_i = \ell_i],
\label{eq:depth_reward}
\end{equation}
yielding a reward of 1 for correct classification and 0 for misclassification. This non-negative formulation ensures training stability in RL while providing a clear binary signal: when $\hat{\ell}_i = 1$, the model triggers elongated reasoning trajectories with reflection on implicit geographical patterns; when $\hat{\ell}_i = 0$, the model proceeds with efficient standard reasoning trajectories.

\subsubsection{Visual Grounding Reward}
\label{sec:grounding_reward}

We introduce a visual grounding objective to suppress hallucinatory generation and ensure the rationale remains strictly tethered to the observed visual elements. This is achieved by evaluating both entity-level accuracy and the overall consistency of the inference path. Formally, given an instance $(I_i, r_i)$, with $\hat{r}_i$ representing the generated rationale and $r_i$ serving as the baseline, we employ named entity recognition to identify the respective sets of distinct entities: $E_i = \{e_1, \dots, e_n\}$ from the prediction and $E_i^{*} = \{e_1^{*}, \dots, e_m^{*}\}$ from the ground truth.

For each predicted entity $e_j \in E_i$, we compute a grounding score $G(e_j, I_i) \in [0, 1]$ using Grounding-DINO \cite{liu2024grounding}, which returns the maximum detection confidence for $e_j$ in the image $I_i$. We define the visual grounding reward as:
\begin{equation}
\begin{split}
R_{vis}(I_i, \hat{r}_i, r_i) = & \underbrace{\frac{1}{|E_i|} \sum_{j=1}^{|E_i|} G(e_j, I_i)}_{R_{grounding}} \cdot \underbrace{\text{J}(E_i, E_i^{*})}_{R_{alignment}} \\
\end{split}
\label{eq:vis_reward}
\end{equation}
where $R_{grounding}$ measures the average grounding confidence of predicted entities, $R_{alignment} = \frac{|E_i \cap E_i^{*}|}{|E_i \cup E_i^{*}|}$ evaluates semantic alignment with the reference trajectory. 

\subsubsection{Hierarchical Geo-localization Reward}
\label{sec:hierarchical_geo_reward}

Image geo-localization predictions exhibit a natural hierarchy (country $\rightarrow$ city $\rightarrow$ coordinates), where coarse-level errors are amplified, leading to larger coordinate inaccuracies.
To address this hierarchical structure, we design a unified well-structured hierarchical reward.

Let the ground-truth location be $g_i = (c_i, t_i, coord_i)$, where $c_i$ and $t_i$ denote the country and city, respectively, and $coord_i$ denotes the coordinates. Let the prediction be $\hat{g}_i = (\hat{c}_i, \hat{t}_i, \widehat{coord}_i)$. Based on the hierarchy inherent to geo-localization, we define a three-tier reward structure:
\begin{equation}
R_{geo}(\hat{g}_i, g_i) =
\begin{cases}
0 & \text{if } \hat{c}_i \neq c_i \quad \\
\lambda_1 \cdot R_{coord} & \text{if } \hat{c}_i = c_i \land \hat{t}_i \neq t_i \quad \\
\lambda_1 + \lambda_2 \cdot R_{coord} & \text{if } \hat{c}_i = c_i \land \hat{t}_i = t_i \quad
\end{cases},
\label{eq:hierarchical_geo_reward}
\end{equation}
where $\lambda_1$ and $\lambda_2$ are hyperparameters satisfying $\lambda_1 + \lambda_2 = 1$, and the continuous coordinate reward $R_{coord} \in [0, 1]$ is computed via distance-based exponential decay:
\begin{equation}
R_{coord} = \exp\left(-\frac{d}{\sigma}\right),
\label{eq:coord_reward}
\end{equation}
with $d$ representing the geodesic distance in kilometers and $\sigma$ as a distance scale parameter. This formulation is logically hierarchical:
(1) A catastrophic error (wrong country) receives zero reward.
(2) A correct country but wrong city receives a reward scaled only by coordinate accuracy.
(3) Only a correct country and city prediction receives the base reward plus the full scaled coordinate reward. 

\subsection{GRPO-Based RL}
\label{sec:grpo_rl}

Using the rewards defined in Section \ref{sec:reward_construction}, we fine-tune the base model with a two-stage curriculum based on GRPO \cite{shao2024deepseekmath}. GRPO trains a policy while maintaining a reference policy for stability. For each prompt, the model samples several candidate answers and scores them, centering rewards within the same prompt to emphasize relative quality. The policy is then updated with a clipped objective and a KL penalty to keep it close to the reference model, ensuring stable, preference-aware learning.
This curriculum splits training into two stages: (1) reasoning formation, where the model learns adaptive reasoning depth and consistent visual grounding, and (2) geo-localization refinement, where geo-prediction accuracy is optimized. The reference policy $\pi_{\text{ref}}$ is initialized from the SFT model to provide a stable starting point. The following sections detail these two stages and their customized reward functions.

\subsubsection{Stage 1: Reasoning Formation}
\label{sec:stage1_formation}

    The first stage trains the model to reason for image geo-localization by optimizing two key skills: (1) adaptive depth: deciding when to reason deeply or briefly, and (2) grounded reasoning: keeping steps faithful to visual cues. We instantiate the GRPO reward for Stage 1 by defining the reward $r_i^{(j)} = R_{\text{stage1}}$ as below:
\begin{equation}
R_{\text{stage1}}(I_i, \hat{\ell}_i, \hat{r}_i) = w_1 \cdot R_{depth}(I_i, \hat{\ell}_i) + w_2 \cdot R_{vis}(I_i, \hat{r}_i, r_i),
\label{eq:stage1_reward}
\end{equation}

where $R_{depth}$ (Eq. \ref{eq:depth_reward}) guides adaptive depth allocation and $R_{vis}$ (Eq. \ref{eq:vis_reward}) enforces grounding consistency. The weights $w_1$ and $w_2$ control the relative importance of these two objectives.
We train Stage 1 for 3 epochs over the curated dataset in Section \ref{sec:data_curation}, producing checkpoint $\theta_{\text{stage1}}^{*}$.

\subsubsection{Stage 2: Geo-Localization Refinement}
\label{sec:stage2_refinement}

The second stage optimizes geo-localization accuracy while preserving the capabilities from Stage 1. Initialize the policy $\pi_{\theta}^{(2)}=\pi_{\theta_{\text{stage1}}^{*}}$. 
The reward used in stage 2 is:
\begin{equation}
R_{\text{stage2}}(\hat{g}_i, g_i) = R_{geo}(\hat{g}_i, g_i),
\label{eq:stage2_reward}
\end{equation}
where $R_{geo}$ is defined in Eq. \ref{eq:hierarchical_geo_reward}. This objective optimization enables the model to concentrate on improving geo-localization accuracy. Crucially, we also update the reference policy to $\pi_{\text{ref}}^{(2)} = \pi_{\theta_{\text{stage1}}^{*}}$. 
We train Stage 2 for 2 epochs.
\section{Experiments}
\label{sec:4_experiments}

We conduct extensive experiments to evaluate our framework, Geo-ADAPT, and its resulting model, Geo-ADAPT-8B. Our evaluation is designed to demonstrate two key points: (1) that our proposed Optimized Locatability Score $L_{opt}$, curated dataset, customized rewards, and two-stage GRPO curriculum yield a superior, adaptive reasoning policy compared to standard supervised fine-tuning or previous RL methods; and (2) that our final model, Geo-ADAPT-8B, achieves state-of-the-art performance against leading open- and closed-source VLMs on multiple geo-localization benchmarks. We also provide qualitative and quantitative analyses to verify that our adaptive policy effectively reduces hallucinations and improves reasoning quality for image geo-localization. More experimental details are provided in the supplementary materials.

\subsection{Experiments Setup}

\textbf{Datasets.} The curated dataset Geo-ADAPT-51K comprises two subsets: Geo-ADAPT-Standard-35K with 35K samples with standard reasoning trajectories and Geo-ADAPT-Augmented-16K with 16K samples with enriched reasoning trajectories. We split both datasets at an 8.5:1.5 ratio for training and testing. Our resulting model, Geo-ADAPT-8B, is evaluated on both our test split and two public geo-localization benchmarks, IM2GPS3K \cite{hays2008im2gps} and YFCC4K \cite{Thomee16YFCC100M} for comprehensive comparison with baseline methods.

\newcommand{\mcol}[2]{\begin{tabular}{@{}c@{}}#1\\#2\end{tabular}}

\begin{table*}[t]
\centering
\caption{Performance comparison for Geo-ADAPT-8B on IM2GPS3K and YFCC4K with baseline methods. For all metrics, higher is better. The best performances are highlighted in \textbf{bold}, while the best performances among baselines are \underline{underlined}. Change represents the relative improvement $\uparrow$ or decline $\downarrow$ of our method over the best baseline.}
\label{tab:my_results}
\resizebox{\textwidth}{!}{%
\begin{tabular}{lcccccccccc}
\toprule
& \multicolumn{5}{c}{\textbf{IM2GPS3K}} & \multicolumn{5}{c}{\textbf{YFCC4K}} \\
\cmidrule(lr){2-6} \cmidrule(lr){7-11}
\textbf{Methods} & \mcol{Street}{1km} & \mcol{City}{25km} & \mcol{Region}{200km} & \mcol{Country}{750km} & \mcol{Continent}{2500km} & \mcol{Street}{1km} & \mcol{City}{25km} & \mcol{Region}{200km} & \mcol{Country}{750km} & \mcol{Continent}{2500km} \\
\midrule
{[L]kNN, sigma=4 \cite{Vo17}} & 7.5 & 18.3 & 27.7 & 38.7 & 57.2 & 2.4 & 6.2 & 9.8 & 23.9 & 43.1 \\
PlaNet \cite{weyand2016planet} & 7.8 & 25.8 & 33.7 & 48.7 & 63.2 & 6.5 & 14.2 & 22.4 & 36.0 & 56.4 \\
CPlaNet \cite{seo2018cplanet} & 11.3 & 25.7 & 34.7 & 47.7 & 65.3 & 6.6 & 15.4 & 21.6 & 36.9 & 54.5 \\
ISNs \cite{MullerBudack18Geolocation} & 10.1 & 29.2 & 35.5 & 50.5 & 64.5 & 6.7 & 15.7 & 25.1 & 36.7 & 55.1 \\
Translocator \cite{pramanick2022world} & 12.1 & 30.5 & 47.6 & 57.7 & 80.5 & 7.6 & 19.7 & 25.6 & 41.8 & 60.2 \\
GeoDecoder \cite{clark2023we} & 11.8 & 34.0 & 45.6 & 62.4 & 74.8 & 10.9 & 23.5 & 34.9 & 48.9 & 69.5 \\
GeoCLIP \cite{vivanco2023geoclip} & 14.3 & 34.0 & 51.4 & 68.7 & 83.9 & 9.3 & 20.4 & 31.4 & 55.6 & 73.9 \\
IMG2Loc \cite{zhou2024img2loc} & 14.9 & 40.7 & 52.3 & 70.0 & 82.7 & 20.3 & 30.0 & 42.4 & 56.7 & 74.3 \\
PIGEON \cite{haas2024pigeon} & 12.3 & 36.5 & 54.3 & 71.8 & 86.1 & 9.3 & 23.8 & 40.2 & 62.9 & 76.8 \\
G3 \cite{Jia24G3} & 15.4 & 42.1 & 55.1 & 71.8 & 83.9 & 24.9 & 34.9 & 47.2 & 64.0 & 78.6 \\
GeoRanker \cite{jia2025georanker} & \textbf{18.7} & \textbf{45.8} & \underline{60.4} & \underline{76.6} & \underline{88.7} & \textbf{33.8} & \textbf{42.3} & \underline{54.8} & \underline{69.1} & \underline{83.9} \\
GRE \cite{Wang25GRESuite} & 13.0 & 34.4 & 52.0 & 69.6 & 86.1 & 10.5 & 35.5 & 43.6 & 68.5 & 78.8 \\
\midrule
\textbf{Geo-ADAPT-8B (Ours)} & 17.9 & 45.3 & \textbf{62.6} & \textbf{77.9} & \textbf{89.5} & 32.5 & 39.1 & \textbf{55.4} & \textbf{70.8} & \textbf{84.5} \\
Change & $\downarrow 4.3\%$ & $\downarrow 1.1\%$ & $\uparrow 3.6\%$ & $\uparrow 1.7\%$ & $\uparrow 0.9\%$ & $\downarrow 3.8\%$ & $\downarrow 7.6\%$ & $\uparrow 1.1\%$ & $\uparrow 2.5\%$ & $\uparrow 0.7\%$ \\
\bottomrule
\end{tabular}%
}
\end{table*}

\textbf{Evaluation Metrics.} We adopt different evaluation metrics to account for the varying annotation formats across datasets. For the Geo-ADAPT-51K test set, we evaluate country- and city-level naming accuracy. For public benchmarks, following standard practices \cite{vivanco2023geoclip, Vo17, Jia24G3, Li24Georeasoner}, we report the percentage of predictions falling within predefined geodesic distance thresholds (1km, 25km, 200km, 750km, and 2500km) from the ground-truth coordinates.

\textbf{Implementation Details.} We implement Geo-ADAPT based on Qwen3-VL-8B \cite{Yang25Qwen3, Qwen2.5-VL}, a publicly available VLM with strong multimodal understanding capabilities. We first train the model for 1 epoch using SFT as a cold start, then apply the two-stage GRPO curriculum described in Section \ref{sec:grpo_rl} with our customized task-specific rewards. All training is conducted on the Geo-ADAPT-51K training split using 4 NVIDIA H200 GPUs on our lab computing server.

\subsection{Main Results}

As shown in Table \ref{tab:my_results}, Geo-ADAPT achieves highly competitive results against state-of-the-art methods on public benchmarks, establishing new SOTA performance on the majority of coarse-grained evaluation thresholds. For instance, on the IM2GPS3K dataset, Geo-ADAPT improves the Region-level (200km) accuracy by 3.6\% and Country-level (750km) accuracy by 1.7\% over the strongest baseline, GeoRanker \cite{jia2025georanker}. This pattern of superior performance on broad-scale metrics is consistent on YFCC4K, with relative gains of 1.1\% for Region-level, 2.5\% for Country-level, and 0.7\% for Continent-level (2500km). A clear trade-off is visible between our reasoning-driven framework and RAG methods like GeoRanker \cite{jia2025georanker} and G3 \cite{Jia24G3}. The RAG baselines excel at fine-grained (Street/City) levels, which is expected as they leverage large retrieval databases for precise instance-level matching. In contrast, Geo-ADAPT operates without an external database at inference time, only relying on its prior knowledge. This distinction explains why our model is outperformed on fine-grained metrics like Street 1km, yet it remains highly comparable at the City level and significantly outperforms these SOTA RAG methods on all coarse-grained metrics across both datasets. 

\begin{table}[t]
\centering
\caption{Localization accuracy for city and country name prediction on the Geo-ADAPT-51K test split compared with baseline reasoning VLMs. $^\dagger$ denotes models that are not publicly available.}
\label{tab:main_results_2}
\resizebox{\columnwidth}{!}{%
\begin{tabular}{lcc}
\toprule
& \multicolumn{2}{c}{\textbf{Geo-ADAPT-51K Test}} \\
\cmidrule(lr){2-3}
\textbf{Methods} & City Name Acc. & Country Name Acc. \\
\midrule
Gemma3-27B \cite{GemmaTeam25Gemma3} & 37.6 & 71.9 \\
InternVL3.5-38B \cite{wang2025internvl3_5} & 36.1 & 79.4 \\
Gemini 2.5 Flash$^\dagger$ \cite{Comanici25Gemini2.5} & \underline{54.1} & \underline{87.2} \\
Qwen3-VL-30B \cite{Qwen2.5-VL} & 43.9 & 83.5 \\
GeoReasoner \cite{Li24Georeasoner} & 31.8 & 67.9 \\
GRE \cite{Wang25GRESuite} & 49.7 & 82.3 \\
\midrule
\textbf{Geo-ADAPT-8B (Ours)} & \textbf{55.8} & \textbf{89.2} \\
Improvement & $\uparrow 3.1\%$ & $\uparrow 2.3\%$ \\
\bottomrule
\end{tabular}
}
\end{table}

Furthermore, to evaluate the model's core reasoning capabilities, we compare it against other leading reasoning VLMs on our Geo-ADAPT-51K test split in Table \ref{tab:main_results_2}. This test measures semantic accuracy (Country/City Name) rather than coordinate distance. The results confirm our framework's superiority: Geo-ADAPT-8B achieves a City Name Accuracy of 55.8 and a Country Name Accuracy of 89.2, outperforming all other open- and closed-source baseline models in our experiments. This represents a relative improvement of 3.1\% (City Name) and 2.3\% (Country Name) over the strongest baseline, Gemini 2.5 Flash$^\dagger$. This combined analysis validates our hypothesis. While RAG methods lead in fine-grained retrieval, Geo-ADAPT demonstrates a superior hierarchical understanding of geography and spatial relationships, enabling it to win on broad-scale distance metrics. More importantly, it also achieves state-of-the-art performance in pure, retrieval-free semantic reasoning, confirming that our locatability-guided adaptive policy learns to reason more accurately and effectively.

\subsection{Ablation Study}
To better understand the contribution and effectiveness of each component, we conduct ablation studies by systematically analyzing key modules of our approach, Geo-ADAPT. (1) w/o $\mathcal{D}_{\text{aug}}$ denotes our dataset without augmented reasoning trajectories with implicit cues. (2) w/o $R_{depth}$ denotes our method without the adaptive reasoning depth reward. (3) w/o $R_{vis}$ denotes our method without the visual grounding reward. (4) w/o $R_{geo}$ denotes our method without the hierarchical geo-localization reward. (5) w/o $\mathcal{T}_{\text{SFT}}$ denotes our method without the SFT cold start.

\newcommand{\mcolL}[2]{\begin{tabular}{@{}l@{}}#1\\#2\end{tabular}}

\begin{table}[h!]
\centering
\caption{Ablation study on IM2GPS3K. The worst performance results in ablation study are \underline{underlined}.}
\label{tab:ablation_study_1}
\resizebox{\columnwidth}{!}{%
\begin{tabular}{lccccc}
\toprule
& \multicolumn{5}{c}{\textbf{IM2GPS3K}} \\
\cmidrule(lr){2-6}
\textbf{Methods} & \mcol{Street}{1km} & \mcol{City}{25km} & \mcol{Region}{200km} & \mcol{Country}{750km} & \mcol{Continent}{2500km} \\
\midrule
w/o $\mathcal{D}_{\text{aug}}$ & 17.7 & 43.2 & 60.1 & \underline{75.4} & 89.2 \\
w/o $R_{depth}$ & 17.2 & 44.3 & 60.5 & 76.7 & 88.9 \\
w/o $R_{vis}$ & 17.4 & 43.7 & 60.8 & 76.8 & 89.0 \\
w/o $R_{geo}$ & \underline{15.3} & \underline{40.4} & \underline{58.9} & 75.7 & \underline{87.9} \\
w/o $\mathcal{T}_{\text{SFT}}$ & 17.8 & 44.9 & 61.3 & 76.5 & 89.1 \\
\midrule
\textbf{Ours} & \textbf{17.9} & \textbf{45.3} & \textbf{62.6} & \textbf{77.9} & \textbf{89.5} \\
\bottomrule
\end{tabular}
}
\end{table}

\begin{table}[h!]
\centering
\caption{Ablation study on Geo-ADAPT-51K test. The worst performance results in ablation study are \underline{underlined}.}
\label{tab:ablation_study_2}
\begin{tabular}{lcc}
\toprule
& \multicolumn{2}{c}{\textbf{Geo-ADAPT-51K Test}} \\
\cmidrule(lr){2-3}
\textbf{Methods} & City Name Acc. & Country Name Acc. \\
\midrule
w/o $\mathcal{D}_{\text{aug}}$ & 53.1 & 87.5 \\
w/o $R_{depth}$ & 54.7 & 88.6 \\
w/o $R_{vis}$ & 53.9 & 87.8 \\
w/o $R_{geo}$ & \underline{51.4} & \underline{86.3} \\
w/o $\mathcal{T}_{\text{SFT}}$ & 54.9 & 88.4 \\
\midrule
\textbf{Ours} & \textbf{55.8} & \textbf{89.2} \\
\bottomrule
\end{tabular}
\end{table}

Tables \ref{tab:ablation_study_1} and \ref{tab:ablation_study_2} present the ablation results on the IM2GPS3K and Geo-ADAPT-51K test, respectively. From these results, we draw several key insights: (1) All components contribute positively to the final performance. The full Geo-ADAPT-8B model consistently outperforms all ablation variants across both datasets, demonstrating the effectiveness of our overall framework design. (2) The hierarchical geo-localization reward ($R_{geo}$) is the most critical component. Removing it results in the most significant performance degradation across all metrics on both datasets. This critically highlights that the Stage 2 geo-localization refinement, which optimizes for localization accuracy with the customized hierarchical geo-localization reward ($R_{geo}$), is essential for achieving SOTA performance. (3) The core components of our adaptive policy are highly effective. Removing either the augmented reasoning trajectories ($\mathcal{D}_{\text{aug}}$) or the adaptive reasoning depth reward ($R_{depth}$) leads to a clear drop in performance. This confirms our hypothesis that explicitly training the model on locatability-stratified data, using a reward that encourages an adaptive policy for reasoning, is superior to a standard fixed-depth approach. (4) Finally, removing either the visual grounding reward ($R_{vis}$) or the SFT cold-start ($\mathcal{T}_{\text{SFT}}$) also degrades performance. This indicates that the SFT initialization provides a strong foundation for the two-stage GRPO curriculum, and that the visual grounding reward ($R_{vis}$) is effective in aligning the model's reasoning with visual evidence, supporting our goal of reducing hallucination.

\section{Conclusion}
\label{sec:5_conclusion}

To address the limitations of fixed-depth reasoning policies in VLM-based geo-localization, we propose Geo-ADAPT, a framework for learning adaptive reasoning policies. We introduce an Optimized Locatability Score $L_{opt}$ to quantify an image's suitability for deep reasoning and use this metric to curate Geo-ADAPT-51K, a novel locatability-stratified dataset. Building on this foundation, we develop a two-stage GRPO curriculum with customized rewards governing adaptive depth, visual grounding, and hierarchical accuracy. Extensive experiments demonstrate that Geo-ADAPT achieves state-of-the-art performance across multiple benchmarks while substantially reducing hallucinations.

\textbf{Limitations.} Despite its effectiveness, Geo-ADAPT faces challenges in fine-grained street-level localization compared to SOTA RAG methods. Without retrieval-based reference candidates, our reasoning-only approach struggles to achieve precise predictions for images lacking distinctive landmarks or readable signage. This limitation reflects a fundamental constraint of current VLMs: the difficulty of inferring exact coordinates from subtle visual cues in a single image without additional information support.

\textbf{Future Directions.} Several promising directions emerge naturally from this work. First, hybrid architectures that dynamically integrate retrieval and deep reasoning based on locatability could effectively combine the strengths of both paradigms. Second, incorporating multi-image reasoning or temporal context from image sequences may further enhance fine-grained localization. Third, extending our adaptive framework to agentic AI systems where models can autonomously select and invoke external tools during their reasoning could enable even more flexible and powerful geo-localization capabilities. Finally, integrating complementary data sources such as remote sensing imagery, satellite views, or cross-view datasets (e.g., street-to-aerial matching) could bridge the gap in fine-grained localization by providing critical additional spatial context that pure reasoning alone simply cannot capture.

{
    \small
    \bibliographystyle{ieeenat_fullname}
    \bibliography{main}
}

\clearpage
\appendix

\section*{Appendix}

\section{Implementation Details}
\label{sec:/1_Implementation_Details}

\subsection{More Information on Training and Inference}

In this section, we provide additional details regarding the training and inference setup. Table \ref{tab:training_details} summarizes the key settings used during these phases. All experiments were conducted on four NVIDIA H200 GPUs. We employed a global batch size of 64 (16 per device) using DeepSpeed Stage 2. In this configuration, the training process consumed approximately 90 GB of GPU memory per device with gradient checkpointing disabled.

\begin{table}[htbp]
    \centering
    \caption{More Details on Training and Inference.}
    \label{tab:training_details}
    \begin{tabular}{ll} 
        \toprule
        \textbf{Parameter} & \textbf{Setting} \\
        \midrule
        GPU & NVIDIA H200 $\times$ 4 \\
        Dataset Samples & 51K \\
        Batch Size & 64 \\
        Batch Size per Device & 16 \\
        Training GPU Memory & 90 GB / GPU \\
        Base Model & Qwen3-VL-8B-Instruct \\
        Deepspeed & Stage 2 \\
        \bottomrule
    \end{tabular}
\end{table}

\subsection{Hyper-parameter Settings}

\begin{table}[t]
\centering
\caption{Hyper-parameter Settings of Proposed Geo-ADAPT.}
\label{tab:hyperparameters}
\begin{tabular}{lc}
\toprule
Hyper-parameter & Value \\
\midrule
\multicolumn{2}{l}{\textit{Optimizer Settings}} \\
Optimizer & AdamW \\
Adam $\beta_1$ & 0.9 \\
Adam $\beta_2$ & 0.95 \\
Weight Decay & 0.1 \\
\midrule
\multicolumn{2}{l}{\textit{Learning Rate Schedule}} \\
SFT Learning Rate & 2e-5 \\
Stage 1 Learning Rate & 1e-6 \\ 
Stage 2 Learning Rate & 1e-6 \\ 
LR Scheduler & Cosine \\
Warmup Ratio & 0.01 \\
\midrule
\multicolumn{2}{l}{\textit{Optimized Locatability Score}} \\
$\gamma_1$ ($L_{reason}$ base) & 0.01 \\ 
$\gamma_2$ ($L_{reason}$ gap) & 0.01 \\ 
$\alpha$ ($L_{opt}$ modulation) & 0.6 \\
$\tau_{\text{margin}}$ (km) & 50 \\
\midrule
\multicolumn{2}{l}{\textit{Reward Weights}} \\
$w_1$ (Depth Reward) & 0.5 \\
$w_2$ (Visual Grounding Reward) & 0.5 \\
$\lambda_1$ (Hierarchical Reward) & 0.3 \\
$\lambda_2$ (Hierarchical Reward) & 0.7 \\
$\sigma$ (Distance Scale, km) & 100 \\
\bottomrule
\end{tabular}
\end{table}

Table~\ref{tab:hyperparameters} provides a summary of the hyperparameter settings employed for training Geo-ADAPT-8B. The configuration is organized into four primary modules: optimizer settings, learning rate schedules, locatability score parameters, and reward weighting. Following state-of-the-art practices for VLM fine-tuning, we utilize the AdamW optimizer with a cosine decay scheduler. Crucially, we adopt a differential learning rate strategy, utilizing a standard rate of 2e-5 for the SFT cold start while reducing it to 1e-6 for the GRPO stages to ensure policy stability during reinforcement learning. Specific parameters governing the Optimized Locatability Score (e.g., $\alpha=0.6$, $\gamma_{1,2}=0.01$) and the hierarchical rewards were empirically calibrated through preliminary ablation studies on a held-out validation set. Unless otherwise noted, this configuration is maintained across all experiments to ensure reproducibility.

\subsection{Prompts}

We standardize the input format for VLMs using a unified prompt template to generate reasoning data. This template establishes an ``expert geospatial analyst'' persona and enforces a structured, three-phase reasoning protocol designed to systematically decompose visual evidence processing. Specifically, the prompt guides the model to first conduct an Explicit Cues Scan (Phase 1) for direct identifiers such as landmarks and text, followed by an Implicit Cues Analysis (Phase 2) focusing on environmental context such as vegetation and architectural styles. These observations are then integrated during Phase 3: Synthesis \& Localization to generate a hierarchical prediction that progresses from region to precise coordinates. To guarantee parsability and interpretability, the output is strictly constrained to a specific format utilizing \texttt{<think>} tags for chain-of-thought reasoning and \texttt{<answer>} tags for the final prediction. This standardized configuration is maintained throughout both the fine-tuning curriculum and final evaluation to ensure equitable performance comparisons.

\FloatBarrier 
\vspace{2mm}

\begin{tcolorbox}[
    colback=gray!10, 
    colframe=gray!80, 
    title=\textbf{Unified Prompt Design}, 
    enhanced,
    breakable,
    arc=0mm,
    outer arc=0mm
]
\textbf{\textless System Instruction\textgreater} \\
You are an expert geospatial analyst. Your task is to infer the location depicted in an input image by systematically analyzing visible evidence. Distinguish between \textbf{Explicit Cues} and \textbf{Implicit Cues} to create a hierarchical prediction.

\vspace{2mm}
\textbf{\textless User Prompt\textgreater} \\
Image: \texttt{[INPUT\_IMAGE]}

\textbf{Begin with a concise checklist (3--7 bullets) outlining your planned analysis steps before starting the detailed reasoning phases.}

Analyze this image to determine its location. Follow this structured, three-phase reasoning process:

\vspace{2mm}
\textbf{PHASE 1: EXPLICIT CUES SCAN} \\
Inspect the image for direct identifiers. If present, analyze and cite them clearly and specifically:
\begin{itemize}[leftmargin=*]
    \item \textbf{Landmarks \& Structures:} Global or national landmarks (e.g., Eiffel Tower), distinctive regional architecture (e.g., identifiable castles, temples), or unique buildings (e.g., specific bridges, religious structures).
    \item \textbf{Language \& Text:} Road signs, storefronts, or graffiti. Identify the script (e.g., Arabic, Cyrillic, Latin) and the precise language.
    \item \textbf{Symbolic Signals:} Administrative identifiers (e.g., license plate formats and country codes), currency, or national flags.
\end{itemize}

\textbf{PHASE 2: IMPLICIT CUES ANALYSIS} \\
Examine the scene for contextual or environmental cues. If present, analyze and cite them specifically:
\begin{itemize}[leftmargin=*]
    \item \textbf{Geographical Features:} Landforms (e.g., karst landscapes, salt flats) and vegetation types (e.g., cacti indicate desert regions, birch forests suggest temperate zones, palm trees indicate tropical areas).
    \item \textbf{Architectural Style \& Layout:} Building styles (e.g., Spanish colonial, Neoclassical, earthen structures) and street features (e.g., cobblestone roads, grid layouts, road markings).
    \item \textbf{Social Characteristics:} Clothing or customs (e.g., kimono, sari, traditional plaid) and transportation modes (e.g., rickshaws, gondolas, distinctive bus colors).
    \item \textbf{Climate \& Atmosphere:} Latitude and climate zone inferred from weather phenomena (e.g., aurora borealis, sandstorms, monsoon rain) and lighting conditions.
\end{itemize}

\textbf{PHASE 3: SYNTHESIS \& LOCALIZATION} \\
Integrate both explicit and implicit evidence to progressively narrow the location: Region $\rightarrow$ Country $\rightarrow$ City $\rightarrow$ Coordinates.

\vspace{2mm}
\textbf{Output Format:}

\textbf{1. Reasoning:} \\
Based on the explicit and implicit cues identified and analyzed, systematically deduce the most likely country and city, then estimate the corresponding coordinates. Present this structured reasoning within \texttt{<think>...</think>} tags, articulating your step-by-step process.

\textbf{2. Final Answer:} \\
Present the final answers in \texttt{<answer>...</answer>} tags using the format below:
\begin{itemize}[leftmargin=*]
    \item \textbf{Country:} [Country Name]
    \item \textbf{City:} [City Name]
    \item \textbf{Estimated Coordinates:} [Latitude (to two decimal places), Longitude (to two decimal places)]
\end{itemize}

\textbf{3. Validation \& Constraints:} \\
After presenting your answer, validate whether all explicit and implicit cues have been adequately considered and clearly referenced. If validation fails, briefly self-correct or clarify limitations before finalizing the output.

\vspace{2mm}
\textbf{Strict Adherence Rules:}
\begin{itemize}[leftmargin=*]
    \item Output must be plaintext only (no JSON or Markdown tables).
    \item Return a single country, city, and coordinate set.
    \item If multiple locations appear equally likely, report only the most probable one and explain the tie-breaking logic in the \texttt{<think>} section.
    \item If no location can be determined due to insufficient cues, set each field in \texttt{<answer>} to ``Unknown'' and provide justification in \texttt{<think>}.
\end{itemize}
\end{tcolorbox}

\section{Qualitative Results}
\label{sec:/2_Qualitative_Results}

To validate the effectiveness of our adaptive reasoning policy, we present qualitative comparisons between Geo-ADAPT-8B and a state-of-the-art baseline, Gemini 2.5 Flash. A defining characteristic of Geo-ADAPT is its ability to dynamically allocate computational resources commensurate with scene complexity. As illustrated in these examples, our framework leverages the Optimized Locatability Score ($L_{opt}$) to modulate reasoning depth: expanding inference steps for more challenging scenes while accelerating convergence for distinct landmarks.

\vspace{1mm}

\begin{figure*}[t] 
    \centering
    \includegraphics[width=1\linewidth]{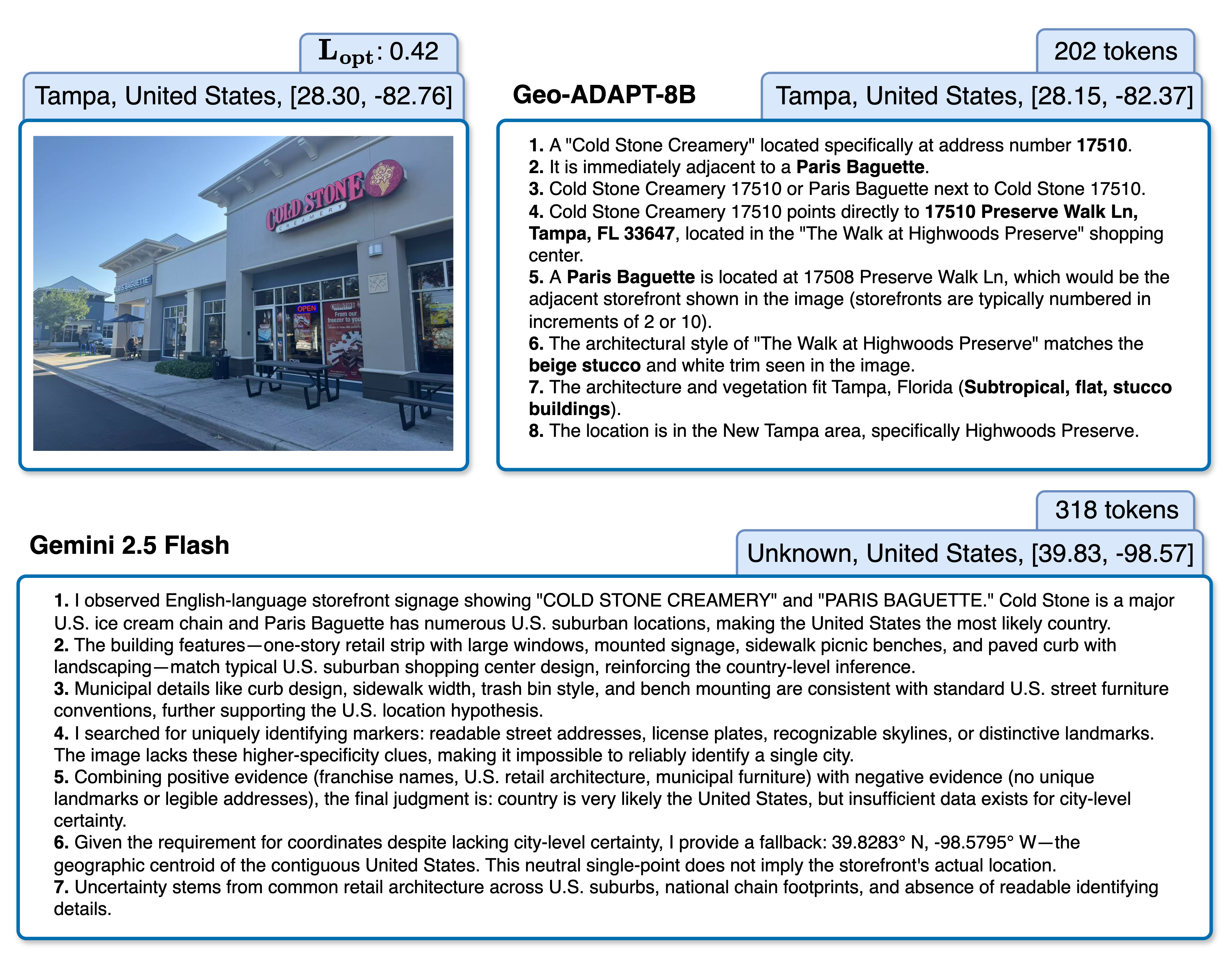}
    \caption{Qualitative comparison on a weakly localizable sample ($L_{opt}=0.42$). Facing visual ambiguity, Geo-ADAPT adaptively expands its reasoning depth (202 tokens), conducting a detailed forensic scan to extract the address ``17510'' and localize the exact shopping center. In contrast, the baseline generates a longer but less grounded response (318 tokens) and fails to identify the city.}
    \label{fig:strip_mall}
\end{figure*}

\noindent\textbf{Deep Reasoning on Weakly Localizable Scenes ($L_{opt}=0.42$).} 
Figure~\ref{fig:strip_mall} demonstrates a challenging scenario characterized by relatively low locatability. The scene depicts a generic suburban strip mall lacking globally recognizable landmarks. Facing this ambiguity, Geo-ADAPT's policy triggers an extended reasoning trajectory, consuming 202 tokens. Guided by the visual grounding reward ($R_{vis}$), the model engages in a multi-step forensic analysis: it explicitly extracts fine-grained textual evidence (the address ``17510''), identifies the specific store combination (``Cold Stone'' and ``Paris Baguette''), and synthesizes these with implicit environmental cues (subtropical vegetation). This deep, deliberative process allows the model to successfully pinpoint the specific shopping center in Tampa, whereas the baseline, despite generating more tokens (318), fails to ground its reasoning in specific evidence and defaults to a generic prediction.

\vspace{1mm}

\begin{figure*}[t]
    \centering
    \includegraphics[width=1\linewidth]{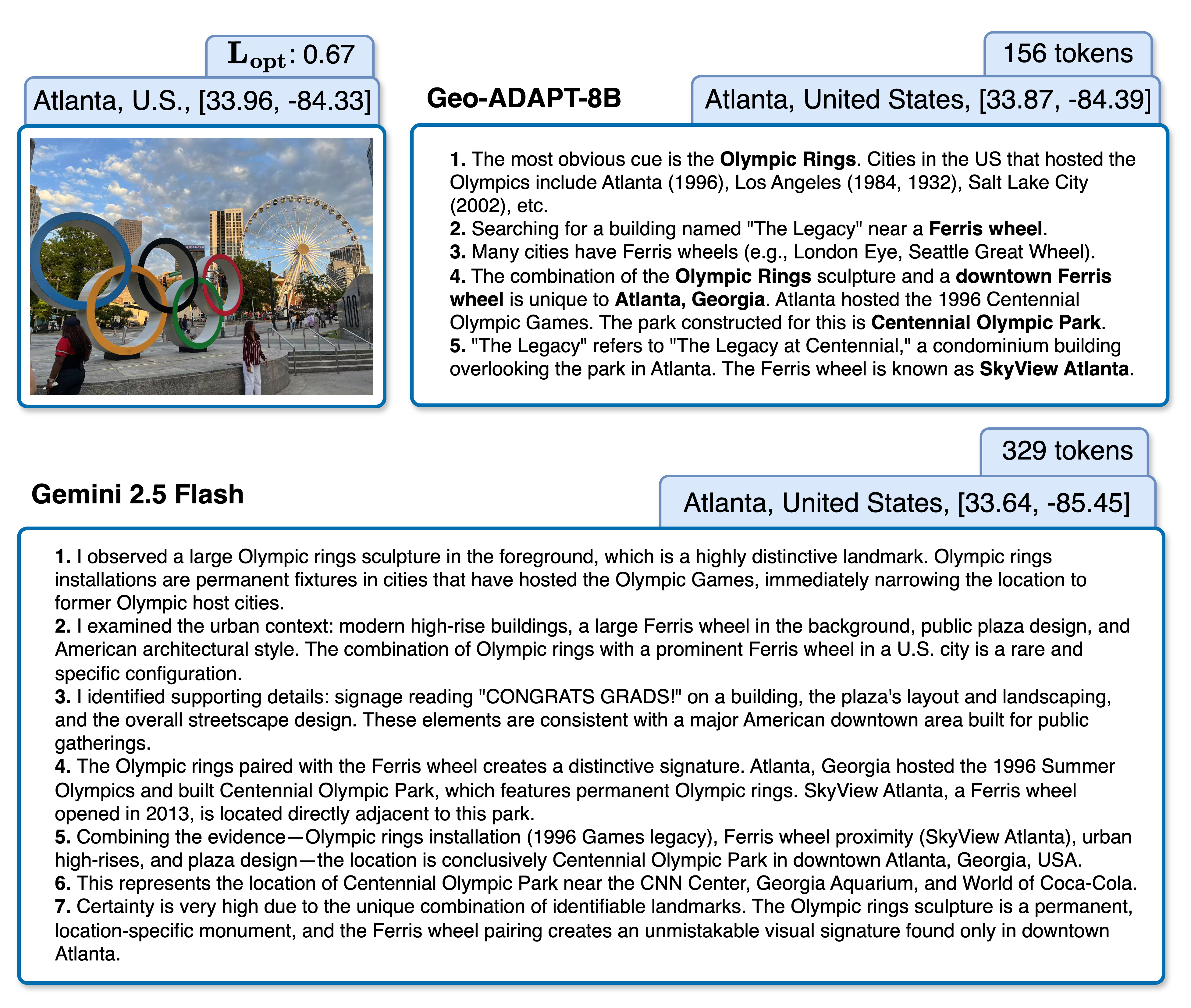}
    \caption{Qualitative comparison on a highly localizable sample ($L_{opt}=0.67$). Recognizing the strong explicit cues (Olympic Rings), Geo-ADAPT adaptively shortens its reasoning path for efficiency (156 tokens), rapidly converging on the correct location. This contrasts with the baseline, which engages in verbose description (329 tokens) without improving precision.}
    \label{fig:atlanta}
\end{figure*}

\noindent\textbf{Efficient Reasoning on Localizable Landmarks ($L_{opt}=0.67$).}
In contrast, Figure~\ref{fig:atlanta} presents a scenario with relatively high locatability containing prominent landmarks. Here, Geo-ADAPT demonstrates adaptive efficiency. Recognizing the high distinctiveness of the Olympic Rings, the policy condenses its reasoning path, requiring only 156 tokens, significantly fewer than the complex case above. Instead of unnecessary exploration, the model rapidly constructs a concise verification chain, cross-referencing the rings with the ``SkyView'' Ferris wheel and ``The Legacy'' building to confirm Atlanta, GA. This demonstrates that our Stage 1 Adaptive Depth Reward ($R_{depth}$) successfully teaches the model to avoid over-reasoning when strong explicit cues are present, achieving precise localization with optimal computational cost.

\end{document}